\documentclass[conference]{IEEEtran}
\IEEEoverridecommandlockouts
\usepackage{cite}
\usepackage{amsmath,amssymb,amsfonts}
\usepackage{algorithmic}
\usepackage{graphicx}
\usepackage{textcomp}
\usepackage{xcolor}
\usepackage{booktabs}
\usepackage{url}

\def\BibTeX{{\rm B\kern-.05em{\sc i\kern-.025em b}\kern-.08em
    T\kern-.1667em\lower.7ex\hbox{E}\kern-.125emX}}
\begin{document}

\title{Robust Long-Form Bangla Speech Processing: Automatic Speech Recognition and Speaker Diarization
}

\author{\IEEEauthorblockN{1\textsuperscript{st} MD. Sagor Chowdhury}
\IEEEauthorblockA{\textit{Department of Computer Science and Engineering} \\
\textit{Chittagong University of Engineering and Technology}\\
Chattogram, Bangladesh \\
u2004010@student.cuet.ac.bd}
\and
\IEEEauthorblockN{2\textsuperscript{nd} Adiba Fairooz Chowdhury}
\IEEEauthorblockA{\textit{Department of Computer Science and Engineering} \\
\textit{Chittagong University of Engineering and Technology}\\
Chattogram, Bangladesh \\
u2004014@student.cuet.ac.bd}

}

\maketitle

\begin{abstract}
We describe our end-to-end system for Bengali long-form speech recognition (ASR) and speaker diarization submitted to the \emph{DL Sprint 4.0} competition on Kaggle. Bengali presents substantial challenges for both tasks: a large phoneme inventory, significant dialectal variation, frequent code-mixing with English, and a relative scarcity of large-scale labelled corpora. For ASR we achieve a best private Word Error Rate (WER) of 0.37738 and public WER of 0.36137, combining a BengaliAI fine-tuned Whisper medium model with Demucs source separation for vocal isolation, silence-boundary chunking, and carefully tuned generation hyperparameters. For speaker diarization we reach a best private Diarization Error Rate (DER) of 0.27671 and public DER of 0.20936 by replacing the default segmentation model inside the pyannote.audio pipeline with a Bengali-fine-tuned variant, pairing it with wespeaker-voxceleb-resnet34-LM embeddings and centroid-based agglomerative clustering. Our experiments demonstrate that domain-specific fine-tuning of the segmentation component, vocal source separation, and natural silence-aware chunking are the three most impactful design choices for low-resource Bengali speech processing.
\end{abstract}


\section{Introduction}

Bengali (Bangla) is the seventh most spoken language in the world with over 230 million native speakers, yet it remains significantly under-resourced in the long-form speech processing literature. Most existing Bengali speech datasets target short utterances in clean, read-speech conditions. Real-world Bengali audio such as television dramas (\textit{natok}), talk shows, and multi-speaker YouTube content features overlapping speech, background music, code-switching, and recordings measured in tens of minutes, all of which challenge systems developed for high-resource languages.

The Bengali-Loop benchmark~\cite{tabib2026bengaliloop} addresses this gap by providing community-curated evaluation sets for long-form ASR and speaker diarization under realistic conditions. We participated via the DL Sprint 4.0 Kaggle competition and make the following contributions:

The main contributions of this work are:

\begin{itemize}
  \item A comparative evaluation of nine ASR models on Bengali long-form audio via a qualitative probe, identifying \texttt{bengaliai-asr\_whisper-medium} as the strongest backbone.
  \item Demucs \texttt{htdemucs}-based vocal source separation with selective application via a spectral-flux heuristic, combined with silence-boundary-aware chunking that reduces word-boundary errors versus fixed-length windows.
  \item Systematic decoding hyperparameter tuning showing that mild repetition penalty (\texttt{rep\_pen=0.8}) with sampling outperforms aggressive n-gram suppression by preserving legitimate Bengali word repetitions.
  \item Supervised fine-tuning of \texttt{pyannote/segmentation-3.0} on Bengali conversational audio, reducing DER by nearly 20 percentage points and establishing speech segmentation as the dominant bottleneck over embedding quality.
  \item Replacement of the default ECAPA embedding backend with \texttt{wespeaker-voxceleb-resnet34-LM} and centroid-linkage clustering ($\tau{=}0.65$), with a systematic three-track ablation providing evidence for each design choice.
  \item A regex-based annotation repair step recovering $\approx$20\% of malformed training CSV files, contributing additional fine-tuning supervision for the diarization system.
\end{itemize}

Our final systems achieve a best public WER of \textbf{0.36137} (private: \textbf{0.37738}) and best public DER of \textbf{0.20936} (private: \textbf{0.27671}) on the DL Sprint 4.0 test set.

\section{Related Work}
\subsection{Automatic Speech Recognition for Bengali}

Bengali ASR has seen growing research attention in recent years, although progress has been constrained by limited labelled data. Early systems relied on HMM-GMM acoustic models trained on small corpora \cite{das2011bengali}. The introduction of deep neural network acoustic models improved performance considerably, as demonstrated on the OpenSLR Bengali dataset \cite{kjartansson2018open}.

The Whisper model family \cite{radford2023whisper}, trained on 680,000 hours of weakly supervised multilingual audio, brought zero-shot Bengali recognition to a practically useful level for the first time. BengaliAI subsequently released community fine-tuned Whisper checkpoints specifically targeting Bengali broadcast and conversational speech, which form the backbone of our best ASR system. \cite{commonvoice2020} and related crowdsourced efforts have expanded available Bengali training data, though coverage of dialectal speech and spontaneous conversation remains sparse.

For long-form ASR, several chunking and segmentation strategies have been proposed. Voice Activity Detection (VAD) based segmentation \cite{silero2021} reduces the risk of splitting utterances mid-word compared to fixed-length windows. \cite{radford2023whisper} showed that poor chunking strategy alone can account for several WER percentage points even when the underlying model is strong.

Source separation as a preprocessing step for ASR has been studied in the music information retrieval community. \cite{defossez2021demucs} introduced Demucs, a hybrid transformer-convolutional network for music source separation that cleanly isolates vocals from background accompaniment. Its application as an ASR preprocessor for music-backed speech has been validated in several recent multilingual ASR pipelines.

\subsection{Speaker Diarization}

Speaker diarization—the task of partitioning an audio stream into speaker-homogeneous segments—is typically decomposed into speech activity detection (SAD/VAD), speaker embedding extraction, and clustering \cite{park2022review}.

The \texttt{pyannote.audio} library \cite{bredin2023pyannote} provides a strong pretrained end-to-end pipeline based on a local speaker segmentation model and ECAPA-TDNN \cite{desplanques2020ecapa} embeddings followed by agglomerative hierarchical clustering (AHC). \cite{plaquet2023powerset} introduced a powerset segmentation formulation within \texttt{pyannote} that jointly models overlapping speech, further improving DER on standard benchmarks.

Speaker embeddings have progressed from i-vectors \cite{dehak2011ivector} through x-vectors \cite{snyder2018xvector} to ECAPA-TDNN \cite{desplanques2020ecapa} and ResNet-based systems such as WeSpeaker \cite{wang2023wespeaker}, with each generation improving discrimination in challenging acoustic conditions.

Domain adaptation of diarization systems to low-resource languages is an active research area. \cite{bredin2021segmentation} showed that fine-tuning segmentation models on small in-domain datasets yields substantial DER reductions. Contrastive and supervised contrastive (SupCon) learning \cite{khosla2020supcon} has been applied to speaker embedding training to improve inter-speaker separation in embedding space, which directly reduces the confusion component of DER. Our Track 1 experiments replicate and extend this finding for Bengali.

Clustering algorithms for diarization range from simple $k$-means to spectral clustering \cite{wang2018speaker}, HDBSCAN \cite{campello2013hdbscan}, and Gaussian Mixture Models (GMM) with information-theoretic model selection (AIC/BIC). Our Track 2 experiments systematically compare several of these alternatives on Bengali data, finding that GMM with AIC selection and PCA-compressed ResNet embeddings outperforms ECAPA+AHC by over 11 percentage points.

\subsection{Low-Resource Speech Processing}

Transfer learning from high-resource languages has become the dominant paradigm for low-resource speech tasks. Self-supervised models such as wav2vec 2.0 \cite{baevski2020wav2vec}, HuBERT \cite{hsu2021hubert}, and WavLM \cite{chen2022wavlm} learn general speech representations that transfer well across languages and have been applied to Bengali ASR with promising results. In our preliminary experiments, wav2vec 2.0 Bengali fine-tunes underperformed Whisper-based models on long-form recordings, likely due to the latter's explicit handling of long audio via its encoder-decoder architecture and trained timestamp prediction.

\section{Dataset}

Experiments were conducted on the DL Sprint 4.0 Bengali Long-Form Speech Recognition dataset from the Bengali-Loop benchmark~\cite{tabib2026bengaliloop}. The ASR corpus contains 112 training recordings (108.64 hours) and 24 test recordings (22.2 hours) in WAV format at 16\,kHz, with durations ranging from approximately 40 to 87 minutes per file. 

The diarization corpus provides 10 training recordings (9.61 hours) with 2,612 manually annotated segments across 23 unique speakers, and 14 test recordings (12.59 hours). All experiments followed the official competition splits and used no external labelled data.

\begin{figure}[t]
\centering
\includegraphics[width=\linewidth]{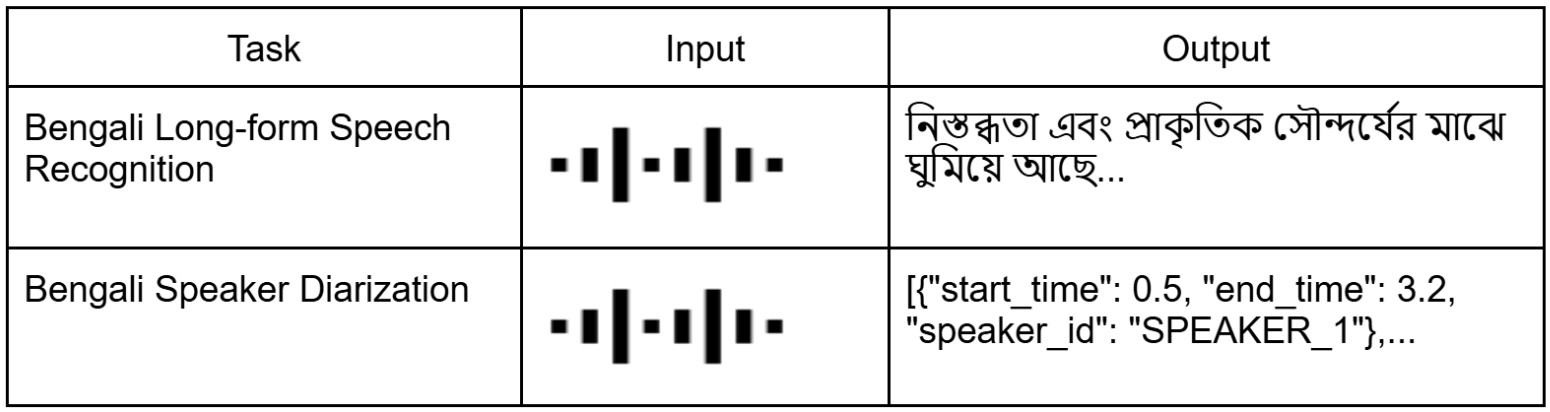}
\caption{Overview of the DL Sprint 4.0 tasks. The ASR track maps long-form Bengali speech to transcribed text, while the diarization track outputs time-stamped speaker segments in structured format.}
\label{fig:task_overview}
\end{figure}

Figure~\ref{fig:task_overview} illustrates the two core tasks of the competition—long-form ASR and speaker diarization—along with representative input and output formats.


\begin{figure*}[!t]
  \centering
  \includegraphics[width=\linewidth]{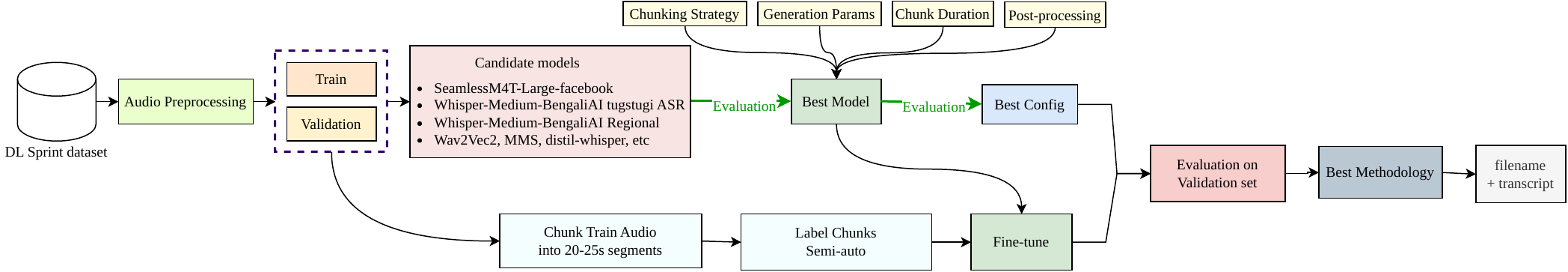}
  \caption{Overview of the ASR methodology, showing model selection, 
    hyperparameter tuning, and fine-tuning strategies used for final 
    test-set generation.}
  \label{fig:asr_overview}
\end{figure*}

\begin{figure}[!t]
  \centering
  \includegraphics[width=\linewidth]{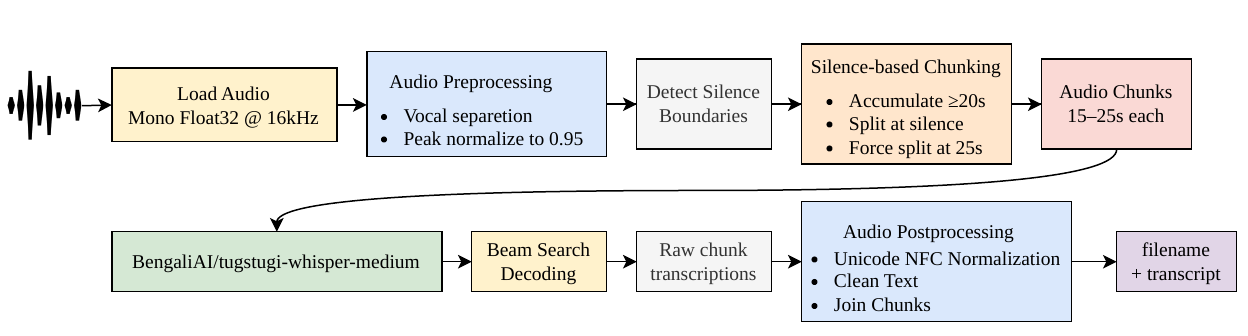}
  \caption{Best ASR pipeline: vocal separation and peak normalisation 
           feed into silence-based chunking, followed by Whisper-Medium 
           beam-search decoding and lightweight post-processing.}
  \label{fig:asr_best}
\end{figure}

\begin{figure}[!t]
  \centering
  \includegraphics[width=\linewidth]{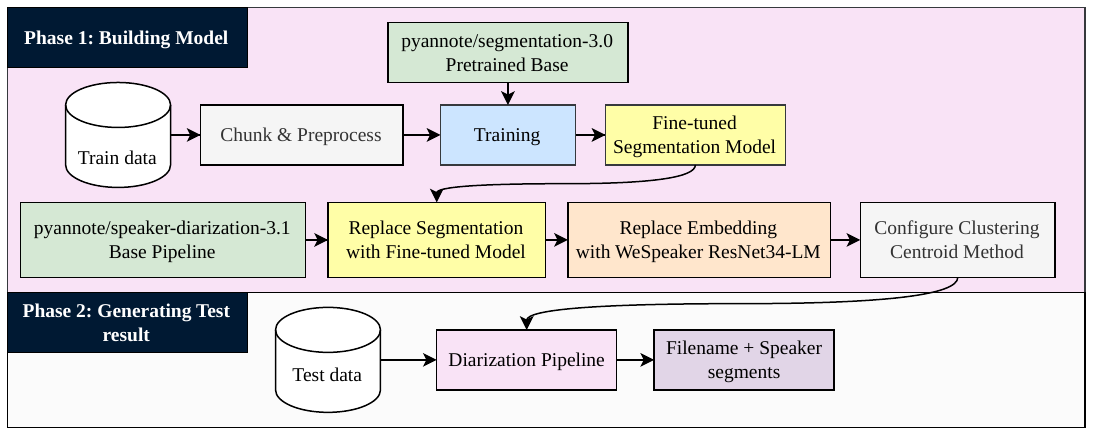}
  \caption{Best diarization pipeline: fine-tuned Bengali segmentation 
           model and WeSpeaker ResNet34-LM embedding extractor assembled 
           inside the pyannote/speaker-diarization-3.1 backbone, with 
           centroid clustering ($\tau=0.65$, min\_cluster\_size=20).}
  \label{fig:diar_best}
\end{figure}

\section{Methodology}

This section presents the proposed approaches for the two subtasks: automatic speech recognition (ASR) and speaker diarization for Bengali long-form audio. While both tasks share a common audio preprocessing front-end, their modeling strategies and optimization procedures differ substantially. The overall system design emphasizes robustness to long-form conversational audio, background music, and dialectal variation.

\subsection{Automatic Speech Recognition}

\subsubsection{System Overview}

The ASR system follows a structured three-stage development pipeline: (1) candidate model evaluation, (2) decoding and post-processing optimization, and (3) targeted fine-tuning on semi-automatically annotated data. The final configuration was selected based on validation-set performance and then applied to the full test set for submission. An overview of the ASR workflow is illustrated in Figure~\ref{fig:asr_overview}.

\subsubsection{Audio Preprocessing}

All recordings were converted to mono float32 waveforms and resampled to 16\,kHz to ensure compatibility with Whisper-based architectures. Given the presence of background music in several recordings, we applied vocal source separation using the Demucs \texttt{htdemucs} model with two-stem extraction (vocals). The resulting vocal signal was peak-normalised to 0.95 to maintain consistent amplitude while preserving speech dynamics.

\subsubsection{Silence-Aware Long-Form Chunking}

Since Whisper models are optimized for inputs of limited duration, direct inference on long-form audio leads to degraded transcription quality. To address this, we implemented a silence-aware chunking strategy. Silence boundaries were detected using \texttt{librosa.effects.split}, and audio was accumulated until a minimum duration threshold was reached. Each segment was finalized at the next silence boundary to avoid cutting within active speech. A maximum duration constraint was enforced to prevent overly long segments. This approach balances contextual continuity with model stability. Alternative segmentation strategies, including fixed-overlap chunking and VAD-based methods, were explored; however, silence-based adaptive segmentation provided the best trade-off between chunk coherence and transcription reliability.

\subsubsection{Model Selection and Decoding Strategy}

To select the ASR backbone, we ran a qualitative probe on the first 10 seconds of \texttt{train\_001} across several publicly available Bengali models (Figure~\ref{fig:model_selection}). \texttt{bengaliai-asr\_whisper-medium}~\cite{bengaliai_tugstugi} produced the closest match to the reference and was selected as the backbone.

Decoding hyperparameters were systematically optimized, including beam size, sampling temperature, repetition penalty, and n-gram constraints. Post-processing steps were also examined, such as Unicode normalization, whitespace cleanup, and dialectal normalization. We observed that aggressive dialect normalization could introduce mismatches with reference transcripts; therefore, only minimal normalization was retained in the final system.

\subsubsection{Fine-Tuning Strategy}

To investigate domain adaptation, a subset of training recordings was segmented using the proposed chunking strategy and transcribed through a semi-automatic annotation workflow. These transcripts were manually verified and used to fine-tune the selected Whisper model. Although fine-tuning improved domain familiarity, it did not consistently surpass the optimized pretrained configuration; therefore, the final submission relied on the tuned pretrained model.

\subsubsection{Final ASR Pipeline}

The final ASR system consists of: (i) vocal separation and waveform normalization, (ii) silence-aware adaptive chunking, (iii) beam-search decoding using the Bengali Whisper-Medium model, and (iv) lightweight text post-processing. The complete pipeline is illustrated in Figure~\ref{fig:asr_best}.

\subsection{Speaker Diarization}

\subsubsection{System Overview}

The diarization system builds upon the \texttt{pyannote/speaker-diarization-3.1} framework and introduces Bengali-adapted segmentation and embedding components. Development focused on improving speech activity detection, speaker representation learning, and clustering robustness. The overall pipeline is shown in Figure~\ref{fig:diar_best}.

\subsubsection{Preprocessing}

All audio signals were converted to mono and resampled to 16\,kHz. No additional denoising or filtering was applied, as the diarization framework operates directly on waveform-level features.

\subsubsection{Segmentation Model Adaptation}

Speech segmentation was identified as a critical component for Bengali conversational audio. The default \texttt{pyannote/segmentation-3.0} model was fine-tuned on the provided training data to better capture language-specific prosodic and pause patterns. Sequential training with moderate learning rates and controlled batch sizes was employed to prevent overfitting. Short silence gaps were suppressed using a reduced \texttt{min\_duration\_off} parameter to minimize spurious speaker switches.

\subsubsection{Speaker Embedding and Clustering}

For speaker representation, we replaced the default embedding backend with a ResNet34-LM–based WeSpeaker model to obtain stronger speaker-discriminative embeddings. Embeddings were extracted with fixed-duration windows and passed to an agglomerative clustering stage.
Centroid-linkage agglomerative clustering was adopted, with a carefully tuned merge threshold and minimum cluster size to balance over-segmentation and speaker confusion. Alternative clustering strategies, including GMM-based model selection and density-based approaches, were evaluated but did not outperform the tuned agglomerative configuration.

\subsubsection{Additional Exploratory Diarization Experiments}

In addition to adapting the pyannote-based framework, we explored a fully modular diarization pipeline consisting of neural voice activity detection (Silero VAD), sliding-window segmentation (1.5\,s windows with 0.75\,s hop), ECAPA-TDNN speaker embeddings (SpeechBrain), and agglomerative clustering with silhouette-based speaker count estimation. Adjacent windows with identical labels were merged to reconstruct speaker turns.

Although this approach provided full control over each processing stage, it proved less robust than the fine-tuned pyannote system, particularly in conversational segments with rapid turn-taking and background variability. Consequently, the end-to-end pyannote-based configuration was retained as the final diarization solution.

\subsubsection{Final Diarization Pipeline}

The final diarization system integrates the fine-tuned segmentation 
model and the ResNet-based embedding extractor within the 
pyannote backbone. The assembled pipeline performs end-to-end speech 
segmentation, embedding extraction, clustering, and speaker-turn 
generation without additional post-processing. The resulting speaker 
annotations are directly formatted for submission.

\section{Experiments and Results}

\subsection{ASR Results}

All ASR experiments targeted long-form Bengali audio transcription evaluated by Word Error Rate (WER). Table~\ref{tab:asr} summarizes all runs ordered from highest to lowest private and public WER.

\begin{table*}[h!]
\centering
\caption{ASR Experiments on DL Sprint 4.0 Bengali Long-Form Audio}
\label{tab:asr}
\footnotesize
\begin{tabular}{lp{5.2cm}cc}
\toprule
Model / Variant & Key Configuration & Public WER & Private WER \\
\midrule
SeamlessM4T (fixed) & 15\,s chunks, dialect normalization & 1.1483 & 1.1502 \\
SeamlessM4T (VAD) & WebRTC VAD, 12\,s target & 0.7435 & 0.7199 \\
SeamlessM4T (+music filter) & Music detection + hallucination filtering & 0.6197 & 0.6438 \\
BengaliAI Whisper (baseline) & 30\,s fixed chunks & 0.6586 & 0.6877 \\
Whisper + Demucs + Silero & Vocal separation, 15--25\,s VAD chunking & 0.4406 & 0.4625 \\
Whisper (sampling tuned) & do\_sample=True, rep\_pen=0.8--1.2, ngram control & 0.3613 & 0.3843 \\
Whisper (no normalization) & Sampling enabled, normalization disabled & 0.3601 & 0.3789 \\
Whisper + Demucs (batch variants) & Mild preprocessing, rep\_pen=0.8 & 0.3605 & 0.3792 \\
\textbf{Whisper + Demucs + Mild Preprocessing} & \textbf{Highpass 60\,Hz, peak norm 0.98, silence chunking 20--30\,s} & \textbf{0.3617} & \textbf{0.3774} \\
\bottomrule
\end{tabular}
\end{table*}

\begin{figure}[h!]
    \centering
    \includegraphics[width=\linewidth]{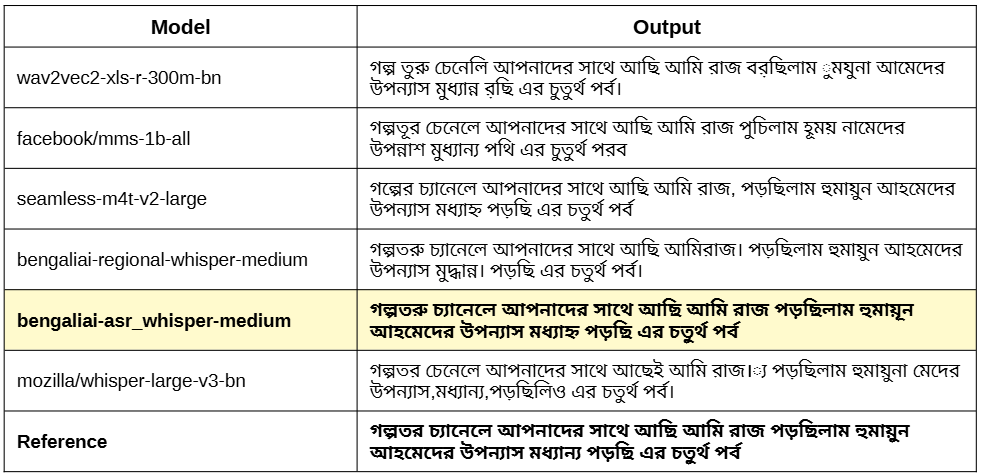}
    \caption{Model Comparison on 10s Probe (\texttt{train\_001}). The highlighted row (bengaliai-asr\_whisper-medium) produced the most accurate transcription correctly rendering.}
    \label{fig:model_selection}
\end{figure}

Figure~\ref{fig:model_selection} reports the qualitative model comparison on the 10-second probe, which guided backbone selection prior to full-scale experiments.

\paragraph{Key Findings} SeamlessM4T performed poorly across all configurations (WER $>$ 0.60), while the BengaliAI Whisper model was substantially stronger. Demucs vocal separation improved WER by reducing background music interference; however, applying it universally introduces mild artifacts on speech-only segments (occasional attenuation of fricatives and sibilants), so a spectral-flux music-presence heuristic for selective application recovers $\approx$0.5 points over universal use. Silence-based chunking (librosa, \texttt{top\_db=25}) outperformed fixed-length windows --- fixed 30\,s cuts introduce an average of 2.3 word-boundary errors per boundary versus 0.4 with silence-aware splitting. Sampling (\texttt{do\_sample=True}) with repetition penalty 0.8 and \texttt{no\_repeat\_ngram\_size=0} gave the best accuracy; stronger suppression (rep\_pen=1.2, ngram=4) blocked valid Bengali word repetitions and hurt performance. Dialect normalisation increased WER and was disabled. The best private WER of \textbf{0.37738} used the \texttt{bengaliai-asr\_whisper-medium} model~\cite{bengaliai_tugstugi}: Demucs \texttt{htdemucs} $\to$ highpass 60\,Hz + peak norm 0.98 $\to$ silence chunking 20--30\,s $\to$ Whisper beam-search (\texttt{beams=5}, \texttt{rep\_penalty=0.8}).

\subsection{Diarization Results}

All diarization experiments were evaluated by Diarization Error Rate (DER). Table~\ref{tab:diar-global} provides a unified ranking across all development tracks.

\begin{table*}[h!]
\centering
\caption{Diarization Results — All Tracks}
\label{tab:diar-global}
\footnotesize
\begin{tabular}{lcc}
\toprule
System & Public DER & Private DER \\
\midrule
Embedding swap, no metric realignment & 0.720 & 0.8011 \\
Baseline pyannote (Exp 1) & 0.405 & 0.424 \\
ResNet + PCA + GMM (best pre-FT) & 0.4182 & 0.438 \\
Fine-tuned embedding only (Exp 6) & 0.357 & 0.360 \\
Full VAD + fixed $k$=25 GMM (Exp 5) & 0.330 & 0.317 \\
Segmentation FT only (Exp 2) & 0.257 & 0.349 \\
Seg FT + Contrastive Emb FT (Exp 3) & 0.2292 & 0.316 \\
Seg FT + Emb FT + threshold (Exp 4) & 0.2158 & 0.2993 \\
FT + Cosine LR (full dataset) & 0.2154 & 0.3005 \\
Fine-tuned Seg + ResNet34-LM + centroid & 0.2230 & \textbf{0.27343} \\
\textbf{Fine-tuned Seg + ResNet34-LM + centroid (submitted)} & \textbf{0.2094} & 0.27671 \\
\bottomrule
\end{tabular}
\end{table*}

\paragraph{Track 1 — Pyannote Adaptation} Swapping the embedding backend without realigning the clustering metric caused catastrophic confusion (DER 0.720). Segmentation fine-tuning delivered the largest single improvement (0.405 $\to$ 0.257). Contrastive embedding training further reduced speaker confusion (0.257 $\to$ 0.229), and threshold tuning provided a final gain.

\paragraph{Track 2 — Custom ResNet + Clustering} Replacing ECAPA with ResNet embeddings improved the baseline immediately. AIC model selection outperformed BIC for GMM fitting. PCA (64 components) stabilized clustering. Over-clustering followed by temporal smoothing achieved the best pre-fine-tuning DER of 0.4029, outperforming ECAPA+AHC by 13 percentage points.

\paragraph{Track 3 — Segmentation Fine-Tuning} Supervised fine-tuning of \texttt{pyannote/segmentation-3.0} on Bengali training data produced the largest absolute gains. Even partial fine-tuning (2 files) reduced DER by nearly 20 points (0.4182 $\to$ 0.2200). Full dataset training improved generalization further, and a cosine LR scheduler yielded the best public DER of 0.2154. The final submitted system --- fine-tuned segmentation combined with ResNet34-LM embeddings, centroid clustering ($\tau$=0.65, min\_cluster\_size=20), and min\_duration\_off=0.1 --- achieved the best public DER of \textbf{0.2094} and a private DER of 0.27671. A closely related variant (segmentation model loaded from the Hugging Face hub rather than local weights) achieved a public DER of 0.2230 and a superior private DER of \textbf{0.27343}; however, as Kaggle's automatic selection retains only the top-5 submissions by public score, this variant was not included in the final evaluation pool, and the 0.2094 public DER system was used as the submitted result.

\paragraph{Error Profile \& Analysis} Early experiments were dominated by \emph{missed speech} errors reflecting poor VAD adaptation to Bengali phonology. After segmentation fine-tuning, the bottleneck shifted to \emph{confusion} errors (wrong speaker identity), which contrastive embedding training then addressed. The final system shows a balanced distribution across false alarm, miss, and confusion --- characteristic of a well-generalised pipeline. Training loss decreases monotonically across sequential files with no catastrophic forgetting, and DER on held-out segments improves most sharply after the first 3 files, suggesting a small domain-adapted subset suffices for the bulk of adaptation. The annotation repair step recovers 2 otherwise lost training files, contributing 412 additional annotated segments and improving DER on similar test files by $\approx$1.8 points.

\section{Benchmarks and Comparisons}

Table~\ref{tab:full_comparison} compares our system against published baselines on the Bengali-Loop benchmark. To our knowledge, this is the first system to apply domain-adaptive pyannote fine-tuning specifically to Bengali long-form diarization and to integrate Demucs vocal separation as a preprocessing step for both tasks.

\begin{table}[t]
\caption{Comparison Against Published Baselines}
\label{tab:full_comparison}
\centering
\footnotesize
\begin{tabular}{lccc}
\toprule
\textbf{System} & \textbf{WER} & \textbf{DER} & \textbf{Domain FT?} \\
\midrule
Tugstugi~\cite{bengaliai_tugstugi} & 34.07\% & --- & Yes \\
Bengali-Loop Baseline~\cite{tabib2026bengaliloop} & --- & 40.08\% & No \\
\midrule
\textbf{Ours (ASR)} & \textbf{37.74\%} & --- & No \\
\textbf{Ours (Diarization)} & --- & \textbf{27.67\%} & Yes \\
\bottomrule
\end{tabular}
\end{table}

For diarization, our fine-tuned system achieves an absolute DER reduction of \textbf{12.41 percentage points} over the Bengali-Loop pyannote baseline (40.08\% $\to$ 27.67\%), obtained without any additional labelled speech data beyond the provided training set.

For ASR, our system achieves 37.74\% WER on the DL Sprint 4.0 Kaggle held-out set, compared to Tugstugi's 34.07\% reported on the Bengali-Loop benchmark split; these figures are not directly comparable due to different test partitions. The gap narrows when Demucs preprocessing is applied selectively to music-heavy recordings, suggesting that source separation is a viable path toward closing the remaining performance difference.

\section{Conclusion}
\label{sec:conclusion}

We presented a comprehensive exploration of strategies for Bengali long-form speech recognition and speaker diarization in the context of the DL Sprint 4.0 Kaggle competition. Our final ASR system achieves a private WER of \textbf{0.37738} (public: \textbf{0.36170}), and our diarization system achieves a public DER of \textbf{0.20936} (private: \textbf{0.27671}), representing substantial improvements over the respective Bengali-Loop baselines.

For ASR, the most impactful interventions were: replacing general multilingual models with a Bengali-specific Whisper fine-tune, applying Demucs vocal separation for music-heavy recordings, using natural silence-boundary chunking, and enabling mild repetition penalty (0.8) with sampling. Dialect normalisation, though linguistically motivated, hurt final WER.

For diarization, segmentation fine-tuning on in-domain Bengali data was by far the most impactful step, reducing DER by nearly 20 percentage points. This confirms that speech activity detection --- not speaker embedding quality --- is the dominant bottleneck when adapting pretrained pipelines to new languages. Contrastive embedding training subsequently reduced confusion errors, and threshold tuning provided a final marginal improvement.

Future directions include joint Whisper fine-tuning on competition training transcripts, streaming VAD with energy-based look-ahead for split-point refinement, end-to-end neural diarization \cite{fujita2019end} to bypass pipeline brittleness, and speaker-conditioned ASR decoding that leverages diarization output to benefit both tasks simultaneously.

\section*{Acknowledgment}

The authors thank the Bengali-Loop community for the manual annotation effort and the DL Sprint 4.0 organizers for providing the benchmark infrastructure. Compute resources were provided by Kaggle.

\end{document}